\documentclass[11pt,a4paper]{article}
\usepackage[hyperref]{acl2020}
\usepackage{times}
\usepackage{latexsym}
\usepackage{booktabs}
\usepackage{graphicx}
\usepackage{amsmath}
\usepackage{amsfonts}

\usepackage{microtype}
\usepackage{float}

\aclfinalcopy 
\def\aclpaperid{1186} 

\newcommand\legal{LegalAI}
\newcommand\ljp{C-LJP}
\newcommand\PLM{PLM}
\newcommand\IE{IE}

\newcommand\LIR{LegalIR}
\newcommand\LQA{LQA}

\title{How Does NLP Benefit Legal System: A Summary of Legal Artificial Intelligence}

\author{
Haoxi Zhong\textsuperscript{\rm 1},
Chaojun Xiao\textsuperscript{\rm 1},
Cunchao Tu\textsuperscript{\rm 1},
Tianyang Zhang\textsuperscript{\rm 2},\\
\textbf{
Zhiyuan Liu\textsuperscript{\rm 1}\thanks{~~Corresponding author.},
Maosong Sun\textsuperscript{\rm 1}} \\
\textsuperscript{\rm 1}Department of Computer Science and Technology\\
Institute for Artificial Intelligence, Tsinghua University, Beijing, China\\
Beijing National Research Center for Information Science and Technology, China \\
\textsuperscript{\rm 2}Beijing Powerlaw Intelligent Technology Co., Ltd., China \\
{\small\tt zhonghaoxi@yeah.net, \{xcjthu,tucunchao\}@gmail.com, zty@powerlaw.ai,}\\
{\small\tt \{lzy,sms\}@tsinghua.edu.cn}\\
}

\date{}

\begin{document}
\maketitle
\begin{abstract}
Legal Artificial Intelligence (\legal{}) focuses on applying the technology of artificial intelligence, especially natural language processing, to benefit tasks in the legal domain. In recent years, \legal{} has drawn increasing attention rapidly from both AI researchers and legal professionals, as \legal{} is beneficial to the legal system for liberating legal professionals from a maze of paperwork. 
Legal professionals often think about how to solve tasks from rule-based and symbol-based methods, while NLP researchers concentrate more on data-driven and embedding methods.
In this paper, we describe the history, the current state, and the future directions of research in \legal{}. We illustrate the tasks from the perspectives of legal professionals and NLP researchers and show several representative applications in \legal{}.
We conduct experiments and provide an in-depth analysis of the advantages and disadvantages of existing works to explore possible future directions.
You can find the implementation of our work from \url{https://github.com/thunlp/CLAIM}.
\end{abstract}

\section{Introduction}

\begin{figure*}[ht]
    \centering
    \includegraphics{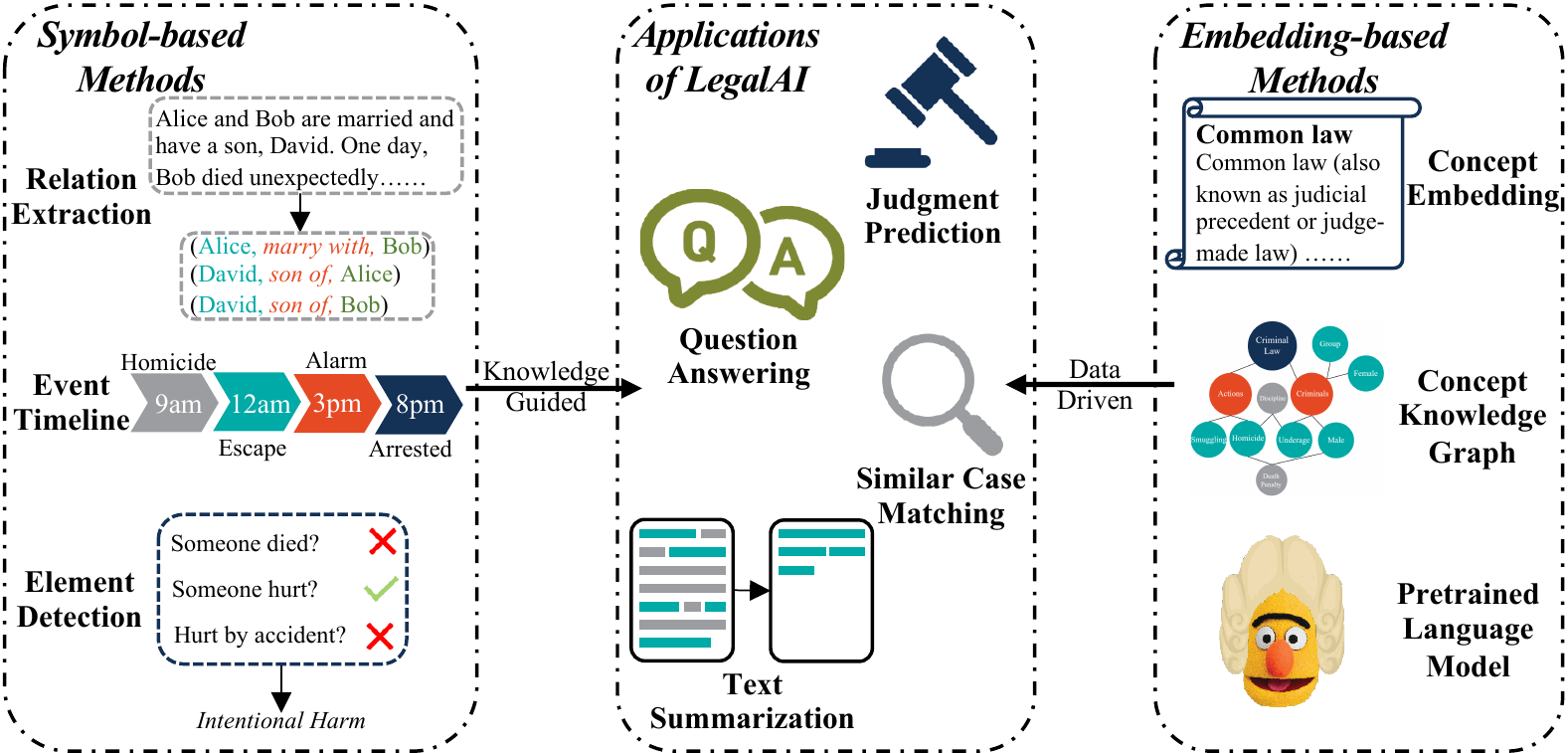}
    \caption{An overview of tasks in \legal{}.}
    \label{fig:overview}
\end{figure*}

Legal Artificial Intelligence (\legal{}) mainly focuses on applying artificial intelligence technology to help legal tasks. The majority of the resources in this field are presented in text forms, such as judgment documents, contracts, and legal opinions. Therefore, most \legal{} tasks are based on Natural Language Processing (NLP) technologies.

\legal{} plays a significant role in the legal domain, as they can reduce heavy and redundant work for legal professionals. Many tasks in the legal domain require the expertise of legal practitioners and a thorough understanding of various legal documents. 
Retrieving and understanding legal documents take lots of time, even for legal professionals. Therefore, a qualified system of \legal{} should reduce the time consumption of these tedious jobs and benefit the legal system. Besides, \legal{} can also provide a reliable reference to those who are not familiar with the legal domain, serving as an affordable form of legal aid.

In order to promote the development of \legal{}, many researchers have devoted considerable efforts over the past few decades. Early works~\cite{kort1957predicting,ulmer1963quantitative,nagel1963applying,segal1984predicting,gardner1984artificial} always use hand-crafted rules or features due to computational limitations at the time. In recent years, with rapid developments in deep learning, researchers begin to apply deep learning techniques to \legal{}. Several new \legal{} datasets have been proposed~\cite{kano2018coliee,xiao2018cail,duan2019cjrc,chalkidis-etal-2019-large,chalkidis-etal-2019-neural}, which can serve as benchmarks for research in the field. Based on these datasets, researchers began exploring NLP-based solutions to a variety of \legal{} tasks, such as Legal Judgment Prediction~\cite{aletras2016predicting,luo-etal-2017-learning-predict,zhong-etal-2018-legal,chen2019charge}, Court View Generation~\cite{ye-etal-2018-interpretable}, Legal Entity Recognition and Classification~\cite{cardellino-etal-2017-legal,angelidis2018named}, Legal Question Answering~\cite{monroy2009nlp,taniguchi2016legal,kim2017two}, Legal Summarization~\cite{hachey2006extractive,bhattacharya2019comparative}.

As previously mentioned, researchers' efforts over the years led to tremendous advances in \legal{}.
To summarize, some efforts concentrate on symbol-based methods, which apply interpretable hand-crafted symbols to legal tasks~\cite{ashley2017artificial,surden2018artificial}. Meanwhile, other efforts with embedding-based methods aim at designing efficient neural models to achieve better performance~\cite{chalkidis2019deep}. 
More specifically, symbol-based methods concentrate more on utilizing interpretable legal knowledge to reason between symbols in legal documents, like events and relationships. Meanwhile, embedding-based methods try to learn latent features for prediction from large-scale data. The differences between these two methods have caused some problems in existing works of \legal{}. Interpretable symbolic models are not effective, and embedding-methods with better performance usually cannot be interpreted, which may bring ethical issues to the legal system such as gender bias and racial discrimination. The shortcomings make it difficult to apply existing methods to real-world legal systems.

We summarize three primary challenges for both embedding-based and symbol-based methods in \legal{}: (1) Knowledge Modelling. Legal texts are well formalized, and there are many domain knowledge and concepts in \legal{}. How to utilize the legal knowledge is of great significance. (2) Legal Reasoning. Although most tasks in NLP require reasoning, the \legal{} tasks are somehow different, as legal reasoning must strictly follow the rules well-defined in law. Thus combining pre-defined rules and AI technology is essential to legal reasoning. Besides, complex case scenarios and complex legal provisions may require more sophisticated reasoning for analyzing. (3) Interpretability. Decisions made in \legal{} usually should be interpretable to be applied to the real legal system. Otherwise, fairness may risk being compromised. Interpretability is as important as performance in \legal{}. 


The main contributions of this work are concluded as follows: 
(1) We describe existing works from the perspectives of both NLP researchers and legal professionals. Moreover, we illustrate several embedding-based and symbol-based methods and explore the future direction of \legal{}. 
(2) We describe three typical applications, including judgment prediction, similar case matching, and legal question answering in detail to emphasize why these two kinds of methods are essential to \legal{}. 
(3) We conduct exhaustive experiments on multiple datasets to explore how to utilize NLP technology and legal knowledge to overcome the challenges in \legal{}. You can find the implementation from github\footnote{\url{https://github.com/thunlp/CLAIM}}.
(4) We summarize \legal{} datasets, which can be regarded as the benchmark for related tasks. The details of these datasets can be found from github\footnote{\url{https://github.com/thunlp/LegalPapers}} with several legal papers worth reading.









\section{Embedding-based Methods}

First, we describe embedding-based methods in \legal{}, also named as representation learning. Embedding-based methods emphasize on representing legal facts and knowledge in embedding space, and they can utilize deep learning methods for corresponding tasks.

\subsection{Character, Word, Concept Embeddings}

Character and word embeddings play a significant role in NLP, as it can embed the discrete texts into continuous vector space. Many embedding methods have been proved effective~\cite{mikolov2013efficient,joulin2016fasttext,pennington2014glove,peters2018deep,yang2014embedding,bordes2013translating,lin2015learning} and they are crucial for the effectiveness of the downstream tasks.

In \legal{}, embedding methods are also essential as they can bridge the gap between texts and vectors. However, it seems impossible to learn the meaning of a professional term directly from some legal factual description. Existing works~\cite{chalkidis2019deep,nay-2016-gov2vec} mainly revolve around applying existing embedding methods like Word2Vec to legal domain corpora. To overcome the difficulty of learning professional vocabulary representations, we can try to capture both grammatical information and legal knowledge in word embedding for corresponding tasks. Knowledge modelling is significant to \legal{}, as many results should be decided according to legal rules and knowledge.

Although knowledge graph methods in the legal domain are promising, there are still two major challenges before their practical usage. Firstly, the construction of the knowledge graph in \legal{} is complicated. In most scenarios, there are no ready-made legal knowledge graphs available, so researchers need to build from scratch. In addition, different legal concepts have different representations and meanings under legal systems in different countries, which also makes it challenging to construct a general legal knowledge graph. Some researchers tried to embed legal dictionaries~\cite{cvrcek-etal-2012-legal}, which can be regarded as an alternative method. Secondly, a generalized legal knowledge graph is different in the form with those commonly used in NLP. Existing knowledge graphs concern the relationship between entities and concepts, but \legal{} focuses more on the explanation of legal concepts. These two challenges make knowledge modelling via embedding in \legal{} non-trivial, and researchers can try to overcome the challenges in the future.

\subsection{Pretrained Language Models}

Pretrained language models (PLMs) such as BERT~\cite{devlin-etal-2019-bert} have been the recent focus in many fields in NLP~\cite{radford2019language,yang2019xlnet,liu2019roberta}. 
Given the success of PLM, using PLM in \legal{} is also a very reasonable and direct choice.
However, there are differences between the text used by existing PLMs and legal text, which also lead to unsatisfactory performances when directly applying PLMs to legal tasks. The differences stem from the terminology and knowledge involved in legal texts. To address this issue, \citet{zhong2019openclap} propose a language model pretrained on Chinese legal documents, including civil and criminal case documents. Legal domain-specific PLMs provide a more qualified baseline system for the tasks of \legal{}. We will show several experiments comparing different BERT models in \legal{} tasks.

For the future exploration of PLMs in \legal{}, researchers can aim more at integrating knowledge into PLMs. Integrating knowledge into pretrained models can help the reasoning ability between legal concepts. Lots of work has been done on integrating knowledge from the general domain into models~\cite{zhang-etal-2019-ernie,peters2019knowledge,hayashi2019latent}. Such technology can also be considered for future application in \legal{}.

\section{Symbol-based Methods}

In this section, we describe symbol-based methods, also named as structured prediction methods. Symbol-based methods are involved in utilizing legal domain symbols and knowledge for the tasks of \legal{}. The symbolic legal knowledge, such as events and relationships, can provide interpretability. Deep learning methods can be employed for symbol-based methods for better performance.

\subsection{Information Extraction}

Information extraction~(\IE{}) has been widely studied in NLP. \IE{} emphasizes on extracting valuable information from texts, and there are many NLP works which concentrate on \IE{}, including name entity recognition~\cite{lample-etal-2016-neural,kuru-etal-2016-charner,akbik-etal-2019-pooled}, relation extraction~\cite{zeng-etal-2015-distant,miwa2016end,lin-etal-2016-neural,christopoulou2018walk}, and event extraction~\cite{chen-etal-2015-event,nguyen-etal-2016-joint,nguyen2018graph}.

\IE{} in \legal{} has also attracted the interests of many researchers. To make better use of the particularity of legal texts, researchers try to use ontology~\cite{bruckschen2010named,cardellino-etal-2017-legal,lenci2009ontology,zhang2017ontological} or global consistency~\cite{yin2018neural} for named entity recognition in \legal{}. To extract relationship and events from legal documents, researchers attempt to apply different NLP technologies, including hand-crafted rules~\cite{bartolini-etal-2004-semantic,truyens-eecke-2014-legal}, 
CRF~\cite{vacek2017sequence}, joint models like SVM, CNN, GRU~\cite{vacek-etal-2019-litigation}, or scale-free identifier network~\cite{yan2017event} for promising results.

Existing works have made lots of efforts to improve the effect of IE, but we need to pay more attention to the benefits of the extracted information. The extracted symbols have a legal basis and can provide interpretability to legal applications, so we cannot just aim at the performance of methods. Here, we show two examples of utilizing the extracted symbols for interpretability of \legal{}:

\textbf{Relation Extraction and Inheritance Dispute}. Inheritance dispute is a type of cases in Civil Law that focuses on the distribution of inheritance rights. Therefore, identifying the relationship between the parties is vital, as those who have the closest relationship with the deceased can get more assets. Towards this goal, relation extraction in inheritance dispute cases can provide the reason for judgment results and improve performance.

\textbf{Event Timeline Extraction and Judgment Prediction of Criminal Case}. In criminal cases, multiple parties are often involved in group crimes. To decide who should be primarily responsible for the crime, we need to determine what everyone has done throughout the case, and the order of these events is also essential. For example, in the case of crowd fighting, the person who fights first should bear the primary responsibility. As a result, a qualified event timeline extraction model is required for judgment prediction of criminal cases.

In future research, we need to concern more about applying extracted information to the tasks of \legal{}. The utilization of such information depends on the requirements of specific tasks, and the information can provide more interpretability.

\subsection{Legal Element Extraction}

In addition to those common symbols in general NLP, \legal{} also has its exclusive symbols, named legal elements. The extraction of legal elements focuses on extracting crucial elements like whether someone is killed or something is stolen. These elements are called constitutive elements of crime, and we can directly convict offenders based on the results of these elements. Utilizing these elements can not only bring intermediate supervision information to the judgment prediction task but also make the prediction results of the model more interpretable.

\begin{table}[ht]
\centering
\small
\begin{tabular} {p{0.85\columnwidth}|l}
\toprule
\multicolumn{2}{p{0.95\columnwidth}}{\textbf{Fact Description:} One day, Bob used a fake reason for marriage decoration to borrow RMB 2k from Alice. After arrested, Bob has paid the money back to Alice.}\\\midrule
Whether did Bob sell something? & $\times$ \\\midrule
Whether did Bob make a fictional fact?  & $\checkmark$\\\midrule
Whether did Bob illegally possess the property of others? & $\checkmark$ \\\midrule
\multicolumn{2}{p{0.95\columnwidth}}{Judgment Results: Fraud.}\\
\bottomrule
\end{tabular}
\caption{An example of element detection from \citet{zhongiteratively}. From this example, we can see that the extracted elements can decide the judgment results. It shows that elements are useful for downstream tasks.}
\label{table:elementexmaple}
\end{table}

Towards a more in-depth analysis of element-based symbols, \citet{element} propose a dataset for extracting elements from three different kinds of cases, including divorce dispute, labor dispute, and loan dispute. The dataset requires us to detect whether the related elements are satisfied or not, and formalize the task as a multi-label classification problem. To show the performance of existing methods on element extraction, we have conducted experiments on the dataset, and the results can be found in Table~\ref{table:element}.

\begin{table}[ht]
    \centering
\resizebox{\columnwidth}{!}{
    \begin{tabular}{c|cc|cc|cc} 
\toprule
 &  \multicolumn{2}{c|}{Divorce} & \multicolumn{2}{c|}{Labor} & \multicolumn{2}{c}{Loan}\\ \midrule
Model & MiF & MaF & MiF & MaF & MiF & MaF \\ \midrule
TextCNN     & 78.7 & 65.9 & 76.4 & 54.4 & 80.3 & 60.6 \\
DPCNN       & 81.3 & 64.0 & 79.8 & 47.4 & 81.4 & 42.5 \\
LSTM        & 80.6 & 67.3 & 81.0 & 52.9 & 80.4 & 53.1 \\
BiDAF       & 83.1 & 68.7 & \textbf{81.5} & \textbf{59.4} & 80.5 & 63.1 \\
BERT     & 83.3 & 69.6 & 76.8 & 43.7 & 78.6 & 39.5 \\
BERT-MS     & \textbf{84.9} & \textbf{72.7} & 79.7 & 54.5 & \textbf{81.9} & \textbf{64.1} \\
\bottomrule
    \end{tabular}
}
\caption{Experimental results on extracting elements. Here MiF and MaF denotes micro-F1 and macro-F1.}
\label{table:element}
\end{table}

We have implemented several classical encoding models in NLP for element extraction, including TextCNN~\cite{kim-2014-convolutional}, DPCNN~\cite{johnson-zhang-2017-deep}, LSTM~\cite{hochreiter1997long}, BiDAF~\cite{seo2016bidirectional}, and BERT~\cite{devlin-etal-2019-bert}. We have tried two different versions of pretrained parameters of BERT, including the origin parameters (BERT) and the parameters pretrained on Chinese legal documents (BERT-MS)~\cite{zhong2019openclap}. From the results, we can see that the language model pretrained on the general domain performs worse than domain-specific \PLM{}, which proves the necessity of \PLM{} in \legal{}. For the following parts of our paper, we will use BERT pretrained on legal documents for better performance.

From the results of element extraction, we can find that existing methods can reach a promising performance on element extraction, but are still not sufficient for corresponding applications. These elements can be regarded as pre-defined legal knowledge and help with downstream tasks. How to improve the performance of element extraction is valuable for further research.

\section{Applications of \legal{}}

In this section, we will describe several typical applications in \legal{}, including Legal Judgment Prediction, Similar Case Matching and Legal Question Answering. Legal Judgment Prediction and Similar Case Matching can be regarded as the core function of judgment in Civil Law and Common Law system, while Legal Question Answering can provide consultancy for those who are unfamiliar with the legal domain. Therefore, exploring these three tasks can cover most aspects of \legal{}.

\subsection{Legal Judgment Prediction}

Legal Judgment Prediction (LJP) is one of the most critical tasks in \legal{}, especially in the Civil Law system. In the Civil Law system, the judgment results are decided according to the facts and the statutory articles. One will receive legal sanctions only after he or she has violated the prohibited acts prescribed by law. The task LJP mainly concerns how to predict the judgment results from both the fact description of a case and the contents of the statutory articles in the Civil Law system.


As a result, LJP is an essential and representative task in countries with Civil Law system like France, Germany, Japan, and China. Besides, LJP has drawn lots of attention from both artificial intelligence researchers and legal professionals. In the following parts, we describe the research progress and explore the future direction of LJP.

\begin{table}[ht]
\centering
\small
\begin{tabular}{l|l}
\toprule
\multicolumn{2}{p{0.95\columnwidth}}{\textbf{Fact Description:} One day, the defendant Bob stole cash 8500 yuan and T-shirts, jackets, pants, shoes, hats (identified a total value of 574.2 yuan) in Beijing Lining store.}\\\midrule
\multicolumn{2}{c}{\textbf{Judgment Results}} \\\midrule
Relevant Articles & Article 264 of Criminal Law. \\\midrule
Applicable Charges & Theft.\\\midrule
Term of Penalty & 6 months.\\
\bottomrule
\end{tabular}
\caption{An example of legal judgment prediction from \citet{zhong-etal-2018-legal}. In this example, the judgment results include relevant articles, applicable charges and the the term of penalty.}
\label{table:ljpexmaple}
\end{table}

\subsubsection*{Related Work}

LJP has a long history. Early works revolve around analyzing existing legal cases in specific circumstances using mathematical or statistical methods~\cite{kort1957predicting,ulmer1963quantitative,nagel1963applying,keown1980mathematical,segal1984predicting,lauderdale2012supreme}. The combination of mathematical methods and legal rules makes the predicted results interpretable.

To promote the progress of LJP, \citet{xiao2018cail} have proposed a large-scale Chinese criminal judgment prediction dataset, \ljp{}. The dataset contains over $2.68$ million legal documents published by the Chinese government, making \ljp{} a qualified benchmark for LJP. \ljp{} contains three subtasks, including relevant articles, applicable charges, and the term of penalty. The first two can be formalized as multi-label classification tasks, while the last one is a regression task. Besides, English LJP datasets also exist~\cite{chalkidis-etal-2019-neural}, but the size is limited.

With the development of the neural network, many researchers begin to explore LJP using deep learning technology~\cite{hu-etal-2018-shot,wang2019using,li2019using,liu2019legal,li2019mann,kang2019creating}. These works can be divided into two primary directions. The first one is to use more novel models to improve performance. \citet{chen2019charge} use the gating mechanism to enhance the performance of predicting the term of penalty. \citet{pan2019charge} propose multi-scale attention to handle the cases with multiple defendants. Besides, other researchers explore how to utilize legal knowledge or the properties of LJP. \citet{luo-etal-2017-learning-predict} use the attention mechanism between facts and law articles to help the prediction of applicable charges. \citet{zhong-etal-2018-legal} present a topological graph to utilize the relationship between different tasks of LJP. Besides, \citet{hu-etal-2018-shot} incorporate ten discriminative legal attributes to help predict low-frequency charges.

\subsubsection*{Experiments and Analysis}
\begin{table*}[ht]
    \centering
    \small
    \begin{tabular}{c|cc|cc|c|cc|cc|c} \toprule
& \multicolumn{5}{c|}{Dev} & \multicolumn{5}{c}{Test} \\\midrule
Task & \multicolumn{2}{c|}{Charge} & \multicolumn{2}{c|}{Article} & Term & \multicolumn{2}{c|}{Charge} & \multicolumn{2}{c|}{Article} & Term \\\midrule
Metrics & MiF & MaF & MiF & MaF & Dis & MiF & MaF & MiF & MaF & Dis \\\midrule
TextCNN
& 93.8 & 74.6 & 92.8 & \textbf{70.5} & 1.586
& 93.9 & 72.2 & 93.5 & 67.0 & 1.539 \\
DPCNN
& 94.7 & 72.2 & 93.9 & 68.8 & 1.448
& 94.9 & 72.1 & 94.6 & 69.4 & 1.390 \\
LSTM
& 94.7 & 71.2 & 93.9 & 66.5 & 1.456
& 94.3 & 66.0 & 94.7 & 70.7 & 1.467 \\
BERT
& 94.5 & 66.3 & 93.5 & 64.7 & \textbf{1.421}
& 94.7 & 71.3 & 94.3 & 66.9 & 1.342 \\\midrule
FactLaw
& 79.5 & 25.4 & 79.8 & 24.9 & 1.721
& 76.9 & 35.0 & 78.1 & 30.8 & 1.683 \\
TopJudge
& \textbf{94.8} & \textbf{76.3} & \textbf{94.0} & 69.6 & 1.438
& \textbf{97.6} & \textbf{76.8} & \textbf{96.9} & \textbf{70.9} & \textbf{1.335} \\
Gating Network
& - & - & - & - & 1.604
& - & - & - & - & 1.553 \\
\bottomrule
    \end{tabular}
    \caption{Experimental results of judgment prediction on C-LJP. In this table, MiF and MaF denotes micro-F1 and macro-F1, and Dis denotes the log distance between prediction and ground truth.}
    \label{table:ljp}
\end{table*}

To better understand recent advances in LJP, we have conducted a series of experiments on \ljp{}. Firstly, we implement several classical text classification models, including TextCNN~\cite{kim-2014-convolutional}, DPCNN~\cite{johnson-zhang-2017-deep}, LSTM~\cite{hochreiter1997long}, and BERT~\cite{devlin-etal-2019-bert}. For the parameters of BERT, we use the pretrained parameters on Chinese criminal cases~\cite{zhong2019openclap}. Secondly, we implement several models which are specially designed for LJP, including FactLaw~\cite{luo-etal-2017-learning-predict}, TopJudge~\cite{zhong-etal-2018-legal}, and Gating Network~\cite{chen2019charge}. The results can be found in Table~\ref{table:ljp}.

From the results, we can learn that most models can reach a promising performance in predicting high-frequency charges or articles. However, the models perform not well on low-frequency labels as there is a gap between micro-F1 and macro-F1. \citet{hu-etal-2018-shot} have explored few-shot learning for LJP. However, their model requires additional attribute information labelled manually, which is time-consuming and makes it hard to employ the model in other datasets.
Besides, we can find that performance of BERT is not satisfactory, as it does not make much improvement from those models with fewer parameters. The main reason is that the length of the legal text is very long, but the maximum length that BERT can handle is $512$. According to statistics, the maximum document length is $56,694$, and the length of $15\%$ documents is over $512$. Document understanding and reasoning techniques are required for LJP.

Although embedding-based methods can achieve promising performance, we still need to consider combining symbol-based with embedding-based methods in LJP. Take TopJudge as an example, this model formalizes topological order between the tasks in LJP (symbol-based part) and uses TextCNN for encoding the fact description. By combining symbol-based and embedding-based methods, TopJudge has achieved promising results on LJP. Comparing the results between TextCNN and TopJudge, we can find that just integrating the order of judgments into the model can lead to improvements, which proves the necessity of combining embedding-based and symbol-based methods.

For better LJP performance, some challenges require the future efforts of researchers:
(1) \textbf{Document understanding and reasoning} techniques are required to obtain global information from extremely long legal texts. 
(2) \textbf{Few-shot learning}. Even low-frequency charges should not be ignored as they are part of legal integrity. Therefore, handling in-frequent labels is essential to LJP.
(3) \textbf{Interpretability}. If we want to apply methods to real legal systems, we must understand how they make predictions. However, existing embedding-based methods work as a black box. What factors affected their predictions remain unknown, and this may introduce unfairness and ethical issues like gender bias to the legal systems. Introducing legal symbols and knowledge mentioned before will benefit the interpretability of LJP.

\subsection{Similar Case Matching}

In those countries with the Common Law system like the United States, Canada, and India, judicial decisions are made according to similar and representative cases in the past. 
As a result, how to identify the most similar case is the primary concern in the judgment of the Common Law system.


In order to better predict the judgment results in the Common Law system, Similar Case Matching (SCM) has become an essential topic of \legal{}. SCM concentrate on finding pairs of similar cases, and the definition of similarity can be various. 
SCM requires to model the relationship between cases from the information of different granularity, like fact level, event level and element level.
In other words, SCM is a particular form of semantic matching \cite{xiao2019cail2019}, which can benefit the legal information retrieval.

\subsubsection*{Related Work}

Traditional methods of Information Retrieve (IR) focus on term-level similarities with statistical models, including TF-IDF~\cite{salton1988term} and BM25~\cite{robertson1994some}, which are widely applied in current search systems. 
In addition to these term matching methods, other researchers try to utilize meta-information~\cite{medin2000psychology,gao2011clickthrough,wu2013learning} to capture semantic similarity. Many machine learning methods have also been applied for IR like SVD~\cite{xu2010relevance} or factorization~\cite{rendle2010factorization,kabbur2013fism}.
With the rapid development of deep learning technology and NLP, many researchers apply neural models, including multi-layer perceptron~\cite{huang2013learning}, CNN~\cite{shen2014latent,hu2014convolutional,qiu2015convolutional}, and RNN~\cite{palangi2016deep} to IR.

There are several \LIR{} datasets, including COLIEE~\cite{kano2018coliee}, CaseLaw~\cite{locke2018test}, and CM~\cite{xiao2019cail2019}. Both COLIEE and CaseLaw are involved in retrieving most relevant articles from a large corpus, while data examples in CM give three legal documents for calculating similarity. These datasets provide benchmarks for the studies of \LIR{}. Many researchers focus on building an easy-to-use legal search engine~\cite{barmakian2000better,turtle1995text}. They also explore utilizing more information, including citations~\cite{monroy2013link,geist2009using,raghav2016analyzing} and legal concepts~\cite{maxwell2008concept,van2017concept}. Towards the goal of calculating similarity in semantic level, deep learning methods have also been applied to \LIR{}. \citet{tran2019building} propose a CNN-based model with document and sentence level pooling which achieves the state-of-the-art results on COLIEE, while other researchers explore employing better embedding methods for \LIR{}~\cite{landthaler2016extending,sugathadasa2018legal}.

\subsubsection*{Experiments and Analysis}

To get a better view of the current progress of \LIR{}, we select CM~\cite{xiao2019cail2019} for experiments. CM contains $8,964$ triples where each triple contains three legal documents $(A,B,C)$. The task designed in CM is to determine whether $B$ or $C$ is more similar to $A$. We have implemented four different types of baselines: (1) Term matching methods, TF-IDF~\cite{salton1988term}. (2) Siamese Network with two parameter-shared encoders, including TextCNN~\cite{kim-2014-convolutional}, BiDAF~\cite{seo2016bidirectional} and BERT~\cite{devlin-etal-2019-bert}, and a distance function. (3) Semantic matching models in sentence level, ABCNN~\cite{yin-etal-2016-abcnn}, and document level, SMASH-RNN~\cite{jiang2019semantic}. The results can be found in Table~\ref{table:scm}.

\begin{table}[h]
    \centering
    \small
    \begin{tabular}{c|c|c} \toprule
         Model & Dev & Test \\\midrule
TF-IDF & 52.9 & 53.3\\\midrule
TextCNN & 62.5 & \textbf{69.9}\\
BiDAF & 63.3 & 68.6 \\
BERT & \textbf{64.3} & 66.8\\\midrule
ABCNN & 62.7 & \textbf{69.9}\\
SMASH RNN & 64.2 &  65.8\\
\bottomrule
    \end{tabular}
    \caption{Experimental results of SCM. The evaluation metric is accuracy.}
    \label{table:scm}
\end{table}

\begin{table*}[ht]
\centering
\small
\begin{tabular}{c|cc|cc|cc}
\toprule
& \multicolumn{2}{c|}{KD-Questions} & \multicolumn{2}{c|}{CA-Questions} & \multicolumn{2}{c}{All} \\ \midrule
                                         & Single               & All                & Single               & All                & Single      & All       \\\midrule
Unskilled Humans        
& $76.9$     & $71.1$     & $62.5$     & $58.0$     & $70.0$     & $64.2$     \\
Skilled Humans          
& $80.6$     & $77.5$     & $86.8$     & $84.7$     & $84.1$     & $81.1$     \\
\midrule
BiDAF       
& $36.7$     & $20.6$     & $37.2$     & $22.2$     & $38.3$     & $22.0$ \\
BERT           
& $\mathbf{38.0}$     & $21.2$     & $38.9$     & $\mathbf{23.7}$     & $39.7$     & $\mathbf{22.3}$ \\
Co-matching     
& $35.8$     & $20.2$     & $35.8$     & $20.3$     & $38.1$     & $21.2$ \\
HAF       
& $36.6$     & $\mathbf{21.4}$     & $\mathbf{42.5}$     & $19.8$     & $\mathbf{42.6}$     & $21.2$ \\
\bottomrule
\end{tabular}
\caption{Experimental results of JEC-QA. The evaluation metrics is accuracy. The performance of unskilled and skilled humans is collected from original paper.}
\label{table:qa}
\end{table*}

From the results, we observe that existing neural models which are capable of capturing semantic information outperform TF-IDF, but the performance is still not enough for SCM. As \citet{xiao2019cail2019} state, the main reason is that legal professionals think that elements in this dataset define the similarity of legal cases. Legal professionals will emphasize on whether two cases have similar elements. Only considering term-level and semantic-level similarity is insufficient for the task. 

For the further study of SCM, there are two directions which need future effort: (1) \textbf{Elemental-based representation}. Researchers can focus more on symbols of legal documents, as the similarity of legal cases is related to these symbols like elements. (2) \textbf{Knowledge incorporation}. As semantic-level matching is insufficient for SCM, we need to consider about incorporating legal knowledge into models to improve the performance and provide interpretability.

\subsection{Legal Question-Answering}

\begin{table}[ht]
\centering
\small
\begin{tabular}{p{0.95\columnwidth}}
\toprule
\textbf{Question}: Which crimes did Alice and Bob commit if they {\color{red}transported more than 1.5 million yuan of counterfeit currency} {\color{blue}from abroad to China}?\\\midrule
\multicolumn{1}{c}{\textbf{Direct Evidence}} \\\midrule
\textbf{P1}: {\color{red}Transportation of counterfeit money}: $\cdots$ The defendants are sentenced to {\color{brown}three years} in prison. \\
\textbf{P2}: {\color{blue}Smuggling counterfeit money}: $\cdots$ The defendants are sentenced to {\color{brown}seven years} in prison. \\ \midrule
\multicolumn{1}{c}{\textbf{Extra Evidence}} \\\midrule
\textbf{P3}: Motivational concurrence: The criminals carry out one behavior but commit several crimes.\\
\textbf{P4}: For motivational concurrence, the criminals should be convicted according to the more serious crime. \\\midrule
\multicolumn{1}{c}{\textbf{Comparison}: {\color{brown}seven years $>$ three years}} \\\midrule
\multicolumn{1}{c}{\textbf{Answer}: Smuggling counterfeit money.}\\
\bottomrule
\end{tabular}
\caption{An example of \LQA{} from \citet{zhong2019jec}. In this example, direct evidence and extra evidence are both required for answering the question. The hard reasoning steps prove the difficulty of legal question answering.}
\label{table:qaexmaple}
\end{table}

Another typical application of \legal{} is Legal Question Answering (\LQA) which aims at answering questions in the legal domain. One of the most important parts of legal professionals' work is to provide reliable and high-quality legal consulting services for non-professionals. However, due to the insufficient number of legal professionals, it is often challenging to ensure that non-professionals can get enough and high-quality consulting services, and LQA is expected to address this issue.

In \LQA{}, the form of questions varies as some questions will emphasize on the explanation of some legal concepts, while others may concern the analysis of specific cases. Besides, questions can also be expressed very differently between professionals and non-professionals, especially when describing domain-specific terms. These problems bring considerable challenges to LQA, and we conduct experiments to demonstrate the difficulties of LQA better in the following parts.

\subsubsection*{Related Work}

In \legal{}, there are many datasets of question answering. \citet{duan2019cjrc} propose CJRC, a legal reading comprehension dataset with the same format as SQUAD 2.0~\cite{rajpurkar-etal-2018-know}, which includes span extraction, yes/no questions, and unanswerable questions. Besides, COLIEE~\cite{kano2018coliee} contains about 500 yes/no questions. Moreover, the bar exam is a professional qualification examination for lawyers, so bar exam datasets~\cite{fawei-etal-2016-passing,zhong2019jec} may be quite hard as they require professional legal knowledge and skills.

In addition to these datasets, researchers have also worked on lots of methods on \LQA{}. The rule-based systems~\cite{buscaldi2010answering,kim2013answering,kim2017two} are prevalent in early research. In order to reach better performance, researchers utilize more information like the explanation of concepts~\cite{taniguchi2016legal,fawei2015using} or formalize relevant documents as graphs to help reasoning~\cite{monroy2009nlp,monroy2008using,tran2013answering}. Machine learning and deep learning methods like CRF~\cite{bach2017question}, SVM~\cite{do2017legal}, and CNN~\cite{kim2015convolutional} have also been applied to \LQA{}. However, most existing methods conduct experiments on small datasets, which makes them not necessarily applicable to massive datasets and real scenarios.

\subsubsection*{Experiments and Analysis}

We select JEC-QA~\cite{zhong2019jec} as the dataset of the experiments, as it is the largest dataset collected from the bar exam, which guarantees its difficulty. JEC-QA contains $28,641$ multiple-choice and multiple-answer questions, together with $79,433$ relevant articles to help to answer the questions. JEC-QA classifies questions into knowledge-driven questions (KD-Questions) and case-analysis questions (CA-Questions) and reports the performances of humans. We implemented several representative question answering models, including BiDAF~\cite{seo2016bidirectional}, BERT~\cite{devlin-etal-2019-bert}, Co-matching~\cite{wang-etal-2018-co}, and HAF~\cite{zhu2018hierarchical}. The experimental results can be found in Table~\ref{table:qa}.

From the experimental results, we can learn the models cannot answer the legal questions well compared with their promising results in open-domain question answering and
there is still a huge gap between existing models and humans in \LQA{}.

For more qualified \LQA{} methods, there are several significant difficulties to overcome: (1) \textbf{Legal multi-hop reasoning}. As \citet{zhong2019jec} state, existing models can perform inference but not multi-hop reasoning. However, legal cases are very complicated, which cannot be handled by single-step reasoning. (2) \textbf{Legal concepts understanding}. We can find that almost all models are better at case analyzing than knowledge understanding, which proves that knowledge modelling is still challenging for existing methods. How to model legal knowledge to LQA is essential as legal knowledge is the foundation of \LQA{}.



\section{Conclusion}

In this paper, we describe the development status of various \legal{} tasks and discuss what we can do in the future. In addition to these applications and tasks we have mentioned, there are many other tasks in \legal{} like legal text summarization and information extraction from legal contracts. Nevertheless, no matter what kind application is, we can apply embedding-based methods for better performance, together with symbol-based methods for more interpretability.

Besides, the three main challenges of legal tasks remain to be solved. Knowledge modelling, legal reasoning, and interpretability are the foundations on which \legal{} can reliably serve the legal domain. Some existing methods are trying to solve these problems, but there is still a long way for researchers to go.

In the future, for these existing tasks, researchers can focus on solving the three most pressing challenges of \legal{} combining embedding-based and symbol-based methods. For tasks that do not yet have a dataset or the datasets are not large enough, we can try to build a large-scale and high-quality dataset or use few-shot or zero-shot methods to solve these problems.

Furthermore, we need to take the \textbf{ethical issues} of \legal{} seriously. Applying the technology of \legal{} directly to the legal system will bring ethical issues like gender bias and racial discrimination. The results given by these methods cannot convince people. To address this issue, we must note that the goal of \legal{} is not replacing the legal professionals but helping their work. As a result, we should regard the results of the models only as a reference. Otherwise, the legal system will no longer be reliable. For example, professionals can spend more time on complex cases and leave the simple cases for the model. However, for safety, these simple cases must still be reviewed. In general, \legal{} should play as a supporting role to help the legal system.

\section*{Acknowledgements}

This work is supported by the National Key Research and Development Program of China (No. 2018YFC0831900) and the National Natural Science Foundation of China (NSFC No. 61772302, 61532010). Besides, the dataset of element extraction is provided by Gridsum.

\bibliography{acl2020}

\begin{thebibliography}{116}
\expandafter\ifx\csname natexlab\endcsname\relax\def\natexlab#1{#1}\fi

\bibitem[{Akbik et~al.(2019)Akbik, Bergmann, and
  Vollgraf}]{akbik-etal-2019-pooled}
Alan Akbik, Tanja Bergmann, and Roland Vollgraf. 2019.
\newblock Pooled contextualized embeddings for named entity recognition.
\newblock In \emph{Proceedings of NAACL}.

\bibitem[{Aletras et~al.(2016)Aletras, Tsarapatsanis, Preotiuc-Pietro, and
  Lampos}]{aletras2016predicting}
Nikolaos Aletras, Dimitrios Tsarapatsanis, Daniel Preotiuc-Pietro, and
  Vasileios Lampos. 2016.
\newblock Predicting judicial decisions of the european court of human rights:
  A natural language processing perspective.
\newblock \emph{PeerJ Computer Science}, 2.

\bibitem[{ANGELIDIS et~al.(2018)ANGELIDIS, CHALKIDIS, and
  KOUBARAKIS}]{angelidis2018named}
Iosif ANGELIDIS, Ilias CHALKIDIS, and Manolis KOUBARAKIS. 2018.
\newblock Named entity recognition, linking and generation for greek
  legislation.

\bibitem[{Ashley(2017)}]{ashley2017artificial}
Kevin~D Ashley. 2017.
\newblock \emph{Artificial intelligence and legal analytics: new tools for law
  practice in the digital age}.
\newblock Cambridge University Press.

\bibitem[{Bach et~al.(2017)Bach, Thien, Phuong et~al.}]{bach2017question}
Ngo~Xuan Bach, Tran Ha~Ngoc Thien, Tu~Minh Phuong, et~al. 2017.
\newblock Question analysis for vietnamese legal question answering.
\newblock In \emph{Proceedings of KSE}. IEEE.

\bibitem[{Barmakian(2000)}]{barmakian2000better}
Deanna Barmakian. 2000.
\newblock Better search engines for law.
\newblock \emph{Law Libr. J.}, 92.

\bibitem[{Bartolini et~al.(2004)Bartolini, Lenci, Montemagni, Pirrelli, and
  Soria}]{bartolini-etal-2004-semantic}
Roberto Bartolini, Alessandro Lenci, Simonetta Montemagni, Vito Pirrelli, and
  Claudia Soria. 2004.
\newblock Semantic mark-up of {I}talian legal texts through {NLP}-based
  techniques.
\newblock In \emph{Proceedings of LREC}.

\bibitem[{Bhattacharya et~al.(2019)Bhattacharya, Hiware, Rajgaria, Pochhi,
  Ghosh, and Ghosh}]{bhattacharya2019comparative}
Paheli Bhattacharya, Kaustubh Hiware, Subham Rajgaria, Nilay Pochhi,
  Kripabandhu Ghosh, and Saptarshi Ghosh. 2019.
\newblock A comparative study of summarization algorithms applied to legal case
  judgments.
\newblock In \emph{Proceedings of ECIR}. Springer.

\bibitem[{Bordes et~al.(2013)Bordes, Usunier, Garcia-Duran, Weston, and
  Yakhnenko}]{bordes2013translating}
Antoine Bordes, Nicolas Usunier, Alberto Garcia-Duran, Jason Weston, and Oksana
  Yakhnenko. 2013.
\newblock Translating embeddings for modeling multi-relational data.
\newblock In \emph{Advances in neural information processing systems}, pages
  2787--2795.

\bibitem[{Bruckschen et~al.(2010)Bruckschen, Northfleet, Bridi, Granada,
  Vieira, Rao, and Sander}]{bruckschen2010named}
M{\'\i}rian Bruckschen, Caio Northfleet, Paulo Bridi, Roger Granada, Renata
  Vieira, Prasad Rao, and Tomas Sander. 2010.
\newblock Named entity recognition in the legal domain for ontology population.
\newblock In \emph{Workshop Programme}, page~16. Citeseer.

\bibitem[{Buscaldi et~al.(2010)Buscaldi, Rosso, G{\'o}mez-Soriano, and
  Sanchis}]{buscaldi2010answering}
Davide Buscaldi, Paolo Rosso, Jos{\'e}~Manuel G{\'o}mez-Soriano, and Emilio
  Sanchis. 2010.
\newblock Answering questions with an n-gram based passage retrieval engine.
\newblock \emph{Journal of Intelligent Information Systems}, 34(2):113--134.

\bibitem[{Cardellino et~al.(2017)Cardellino, Teruel, Alonso~Alemany, and
  Villata}]{cardellino-etal-2017-legal}
Cristian Cardellino, Milagro Teruel, Laura Alonso~Alemany, and Serena Villata.
  2017.
\newblock Legal {NERC} with ontologies, {W}ikipedia and curriculum learning.
\newblock In \emph{Proceedings of EACL}.

\bibitem[{Chalkidis et~al.(2019{\natexlab{a}})Chalkidis, Androutsopoulos, and
  Aletras}]{chalkidis-etal-2019-neural}
Ilias Chalkidis, Ion Androutsopoulos, and Nikolaos Aletras. 2019{\natexlab{a}}.
\newblock Neural legal judgment prediction in {E}nglish.
\newblock In \emph{Proceedings of ACL}.

\bibitem[{Chalkidis et~al.(2019{\natexlab{b}})Chalkidis, Fergadiotis,
  Malakasiotis, and Androutsopoulos}]{chalkidis-etal-2019-large}
Ilias Chalkidis, Emmanouil Fergadiotis, Prodromos Malakasiotis, and Ion
  Androutsopoulos. 2019{\natexlab{b}}.
\newblock Large-scale multi-label text classification on {EU} legislation.
\newblock In \emph{Proceedings of ACL}.

\bibitem[{Chalkidis and Kampas(2019)}]{chalkidis2019deep}
Ilias Chalkidis and Dimitrios Kampas. 2019.
\newblock Deep learning in law: early adaptation and legal word embeddings
  trained on large corpora.
\newblock \emph{Artificial Intelligence and Law}, 27(2):171--198.

\bibitem[{Chen et~al.(2019)Chen, Cai, Dai, Dai, and Ding}]{chen2019charge}
Huajie Chen, Deng Cai, Wei Dai, Zehui Dai, and Yadong Ding. 2019.
\newblock Charge-based prison term prediction with deep gating network.
\newblock In \emph{Proceedings of EMNLP-IJCNLP}, pages 6363--6368.

\bibitem[{Chen et~al.(2015)Chen, Xu, Liu, Zeng, and
  Zhao}]{chen-etal-2015-event}
Yubo Chen, Liheng Xu, Kang Liu, Daojian Zeng, and Jun Zhao. 2015.
\newblock Event extraction via dynamic multi-pooling convolutional neural
  networks.
\newblock In \emph{Proceedings of ACL}.

\bibitem[{Christopoulou et~al.(2018)Christopoulou, Miwa, and
  Ananiadou}]{christopoulou2018walk}
Fenia Christopoulou, Makoto Miwa, and Sophia Ananiadou. 2018.
\newblock A walk-based model on entity graphs for relation extraction.
\newblock In \emph{Proceedings of ACL}, pages 81--88.

\bibitem[{Cvr{\v{c}}ek et~al.(2012)Cvr{\v{c}}ek, Pala, and
  Rychl{\'y}}]{cvrcek-etal-2012-legal}
Franti{\v{s}}ek Cvr{\v{c}}ek, Karel Pala, and Pavel Rychl{\'y}. 2012.
\newblock Legal electronic dictionary for {C}zech.
\newblock In \emph{Proceedings of LREC}.

\bibitem[{Devlin et~al.(2019)Devlin, Chang, Lee, and
  Toutanova}]{devlin-etal-2019-bert}
Jacob Devlin, Ming-Wei Chang, Kenton Lee, and Kristina Toutanova. 2019.
\newblock {BERT}: Pre-training of deep bidirectional transformers for language
  understanding.
\newblock In \emph{Proceedings of NAACL}.

\bibitem[{Do et~al.(2017)Do, Nguyen, Tran, Nguyen, and Nguyen}]{do2017legal}
Phong-Khac Do, Huy-Tien Nguyen, Chien-Xuan Tran, Minh-Tien Nguyen, and Minh-Le
  Nguyen. 2017.
\newblock Legal question answering using ranking svm and deep convolutional
  neural network.
\newblock \emph{arXiv preprint arXiv:1703.05320}.

\bibitem[{Duan et~al.(2019)Duan, Wang, Wang, Ma, Cui, Wu, Wang, Liu, Huo, Hu
  et~al.}]{duan2019cjrc}
Xingyi Duan, Baoxin Wang, Ziyue Wang, Wentao Ma, Yiming Cui, Dayong Wu, Shijin
  Wang, Ting Liu, Tianxiang Huo, Zhen Hu, et~al. 2019.
\newblock Cjrc: A reliable human-annotated benchmark dataset for chinese
  judicial reading comprehension.
\newblock In \emph{Proceedings of CCL}. Springer.

\bibitem[{Fawei et~al.(2016)Fawei, Wyner, and Pan}]{fawei-etal-2016-passing}
Biralatei Fawei, Adam Wyner, and Jeff Pan. 2016.
\newblock Passing a {USA} national bar exam: a first corpus for
  experimentation.
\newblock In \emph{Proceedings of LREC}.

\bibitem[{Fawei et~al.(2015)Fawei, Wyner, Pan, and
  Kollingbaum}]{fawei2015using}
Biralatei Fawei, Adam Wyner, Jeff~Z Pan, and Martin Kollingbaum. 2015.
\newblock Using legal ontologies with rules for legal textual entailment.
\newblock In \emph{AI Approaches to the Complexity of Legal Systems}, pages
  317--324. Springer.

\bibitem[{Gao et~al.(2011)Gao, Toutanova, and Yih}]{gao2011clickthrough}
Jianfeng Gao, Kristina Toutanova, and Wen-tau Yih. 2011.
\newblock Clickthrough-based latent semantic models for web search.
\newblock In \emph{Proceedings of SIGIR}. ACM.

\bibitem[{Gardner(1984)}]{gardner1984artificial}
Anne von der~Lieth Gardner. 1984.
\newblock An artificial intelligence approach to legal reasoning.

\bibitem[{Geist(2009)}]{geist2009using}
Anton Geist. 2009.
\newblock Using citation analysis techniques for computer-assisted legal
  research in continental jurisdictions.
\newblock \emph{Available at SSRN 1397674}.

\bibitem[{Hachey and Grover(2006)}]{hachey2006extractive}
Ben Hachey and Claire Grover. 2006.
\newblock Extractive summarisation of legal texts.
\newblock \emph{Artificial Intelligence and Law}, 14(4):305--345.

\bibitem[{Hayashi et~al.(2019)Hayashi, Hu, Xiong, and
  Neubig}]{hayashi2019latent}
Hiroaki Hayashi, Zecong Hu, Chenyan Xiong, and Graham Neubig. 2019.
\newblock Latent relation language models.
\newblock \emph{arXiv preprint arXiv:1908.07690}.

\bibitem[{Hochreiter and Schmidhuber(1997)}]{hochreiter1997long}
Sepp Hochreiter and J{\"u}rgen Schmidhuber. 1997.
\newblock Long short-term memory.
\newblock \emph{Neural computation}, 9(8).

\bibitem[{Hu et~al.(2014)Hu, Lu, Li, and Chen}]{hu2014convolutional}
Baotian Hu, Zhengdong Lu, Hang Li, and Qingcai Chen. 2014.
\newblock Convolutional neural network architectures for matching natural
  language sentences.
\newblock In \emph{Proceedings of NIPS}.

\bibitem[{Hu et~al.(2018)Hu, Li, Tu, Liu, and Sun}]{hu-etal-2018-shot}
Zikun Hu, Xiang Li, Cunchao Tu, Zhiyuan Liu, and Maosong Sun. 2018.
\newblock Few-shot charge prediction with discriminative legal attributes.
\newblock In \emph{Proceedings of COLING}.

\bibitem[{Huang et~al.(2013)Huang, He, Gao, Deng, Acero, and
  Heck}]{huang2013learning}
Po-Sen Huang, Xiaodong He, Jianfeng Gao, Li~Deng, Alex Acero, and Larry Heck.
  2013.
\newblock Learning deep structured semantic models for web search using
  clickthrough data.
\newblock In \emph{Proceedings of CIKM}. ACM.

\bibitem[{Jiang et~al.(2019)Jiang, Zhang, Li, Bendersky, Golbandi, and
  Najork}]{jiang2019semantic}
Jyun-Yu Jiang, Mingyang Zhang, Cheng Li, Michael Bendersky, Nadav Golbandi, and
  Marc Najork. 2019.
\newblock Semantic text matching for long-form documents.
\newblock In \emph{Proceedings of WWW}. ACM.

\bibitem[{Johnson and Zhang(2017)}]{johnson-zhang-2017-deep}
Rie Johnson and Tong Zhang. 2017.
\newblock Deep pyramid convolutional neural networks for text categorization.
\newblock In \emph{Proceedings of ACL}.

\bibitem[{Joulin et~al.(2016)Joulin, Grave, Bojanowski, Douze, J{\'e}gou, and
  Mikolov}]{joulin2016fasttext}
Armand Joulin, Edouard Grave, Piotr Bojanowski, Matthijs Douze, H{\'e}rve
  J{\'e}gou, and Tomas Mikolov. 2016.
\newblock Fasttext. zip: Compressing text classification models.
\newblock \emph{arXiv preprint arXiv:1612.03651}.

\bibitem[{Kabbur et~al.(2013)Kabbur, Ning, and Karypis}]{kabbur2013fism}
Santosh Kabbur, Xia Ning, and George Karypis. 2013.
\newblock Fism: factored item similarity models for top-n recommender systems.
\newblock In \emph{Proceedings of SIGKDD}. ACM.

\bibitem[{Kang et~al.(2019)Kang, Liu, Liu, Shi, and Ye}]{kang2019creating}
Liangyi Kang, Jie Liu, Lingqiao Liu, Qinfeng Shi, and Dan Ye. 2019.
\newblock Creating auxiliary representations from charge definitions for
  criminal charge prediction.
\newblock \emph{arXiv preprint arXiv:1911.05202}.

\bibitem[{Kano et~al.(2018)Kano, Kim, Yoshioka, Lu, Rabelo, Kiyota, Goebel, and
  Satoh}]{kano2018coliee}
Yoshinobu Kano, Mi-Young Kim, Masaharu Yoshioka, Yao Lu, Juliano Rabelo, Naoki
  Kiyota, Randy Goebel, and Ken Satoh. 2018.
\newblock Coliee-2018: Evaluation of the competition on legal information
  extraction and entailment.
\newblock In \emph{Proceedings of JSAI}, pages 177--192. Springer.

\bibitem[{Keown(1980)}]{keown1980mathematical}
R~Keown. 1980.
\newblock Mathematical models for legal prediction.
\newblock \emph{Computer/LJ}, 2:829.

\bibitem[{Kim and Goebel(2017)}]{kim2017two}
Mi-Young Kim and Randy Goebel. 2017.
\newblock Two-step cascaded textual entailment for legal bar exam question
  answering.
\newblock In \emph{Proceedings of Articial Intelligence and Law}. ACM.

\bibitem[{Kim et~al.(2015)Kim, Xu, and Goebel}]{kim2015convolutional}
Mi-Young Kim, Ying Xu, and Randy Goebel. 2015.
\newblock A convolutional neural network in legal question answering.

\bibitem[{Kim et~al.(2013)Kim, Xu, Goebel, and Satoh}]{kim2013answering}
Mi-Young Kim, Ying Xu, Randy Goebel, and Ken Satoh. 2013.
\newblock Answering yes/no questions in legal bar exams.
\newblock In \emph{Proceedings of JSAI}, pages 199--213. Springer.

\bibitem[{Kim(2014)}]{kim-2014-convolutional}
Yoon Kim. 2014.
\newblock Convolutional neural networks for sentence classification.
\newblock In \emph{Proceedings of EMNLP}.

\bibitem[{Kort(1957)}]{kort1957predicting}
Fred Kort. 1957.
\newblock Predicting supreme court decisions mathematically: A quantitative
  analysis of the "right to counsel" cases.
\newblock \emph{American Political Science Review}, 51(1):1--12.

\bibitem[{Kuru et~al.(2016)Kuru, Can, and Yuret}]{kuru-etal-2016-charner}
Onur Kuru, Ozan~Arkan Can, and Deniz Yuret. 2016.
\newblock {C}har{NER}: Character-level named entity recognition.
\newblock In \emph{Proceedings of COLING}.

\bibitem[{Lample et~al.(2016)Lample, Ballesteros, Subramanian, Kawakami, and
  Dyer}]{lample-etal-2016-neural}
Guillaume Lample, Miguel Ballesteros, Sandeep Subramanian, Kazuya Kawakami, and
  Chris Dyer. 2016.
\newblock Neural architectures for named entity recognition.
\newblock In \emph{Proceedings of NAACL}.

\bibitem[{Landthaler et~al.(2016)Landthaler, Waltl, Holl, and
  Matthes}]{landthaler2016extending}
J{\"o}rg Landthaler, Bernhard Waltl, Patrick Holl, and Florian Matthes. 2016.
\newblock Extending full text search for legal document collections using word
  embeddings.
\newblock In \emph{JURIX}, pages 73--82.

\bibitem[{Lauderdale and Clark(2012)}]{lauderdale2012supreme}
Benjamin~E Lauderdale and Tom~S Clark. 2012.
\newblock The supreme court's many median justices.
\newblock \emph{American Political Science Review}, 106(4):847--866.

\bibitem[{Lenci et~al.(2009)Lenci, Montemagni, Pirrelli, and
  Venturi}]{lenci2009ontology}
Alessandro Lenci, Simonetta Montemagni, Vito Pirrelli, and Giulia Venturi.
  2009.
\newblock Ontology learning from italian legal texts.
\newblock \emph{Law, Ontologies and the Semantic Web}, 188:75--94.

\bibitem[{Li et~al.(2019{\natexlab{a}})Li, Zhang, Ye, Guo, and
  Fang}]{li2019mann}
Shang Li, Hongli Zhang, Lin Ye, Xiaoding Guo, and Binxing Fang.
  2019{\natexlab{a}}.
\newblock Mann: A multichannel attentive neural network for legal judgment
  prediction.
\newblock \emph{IEEE Access}.

\bibitem[{Li et~al.(2019{\natexlab{b}})Li, He, Yan, Zhang, and
  Wang}]{li2019using}
Yu~Li, Tieke He, Ge~Yan, Shu Zhang, and Hui Wang. 2019{\natexlab{b}}.
\newblock Using case facts to predict penalty with deep learning.
\newblock In \emph{International Conference of Pioneering Computer Scientists,
  Engineers and Educators}, pages 610--617. Springer.

\bibitem[{Lin et~al.(2015)Lin, Liu, Sun, Liu, and Zhu}]{lin2015learning}
Yankai Lin, Zhiyuan Liu, Maosong Sun, Yang Liu, and Xuan Zhu. 2015.
\newblock Learning entity and relation embeddings for knowledge graph
  completion.
\newblock In \emph{Proceedings of AAAI}.

\bibitem[{Lin et~al.(2016)Lin, Shen, Liu, Luan, and Sun}]{lin-etal-2016-neural}
Yankai Lin, Shiqi Shen, Zhiyuan Liu, Huanbo Luan, and Maosong Sun. 2016.
\newblock Neural relation extraction with selective attention over instances.
\newblock In \emph{Proceedings of ACL}.

\bibitem[{Liu et~al.(2019{\natexlab{a}})Liu, Ott, Goyal, Du, Joshi, Chen, Levy,
  Lewis, Zettlemoyer, and Stoyanov}]{liu2019roberta}
Yinhan Liu, Myle Ott, Naman Goyal, Jingfei Du, Mandar Joshi, Danqi Chen, Omer
  Levy, Mike Lewis, Luke Zettlemoyer, and Veselin Stoyanov. 2019{\natexlab{a}}.
\newblock Roberta: A robustly optimized bert pretraining approach.
\newblock \emph{arXiv preprint arXiv:1907.11692}.

\bibitem[{Liu et~al.(2019{\natexlab{b}})Liu, Tu, and Sun}]{liu2019legal}
Zhiyuan Liu, Cunchao Tu, and Maosong Sun. 2019{\natexlab{b}}.
\newblock Legal cause prediction with inner descriptions and outer hierarchies.
\newblock In \emph{Proceedings of CCL}, pages 573--586. Springer.

\bibitem[{Locke and Zuccon(2018)}]{locke2018test}
Daniel Locke and Guido Zuccon. 2018.
\newblock A test collection for evaluating legal case law search.
\newblock In \emph{Proceedings of SIGIR}. ACM.

\bibitem[{Luo et~al.(2017)Luo, Feng, Xu, Zhang, and
  Zhao}]{luo-etal-2017-learning-predict}
Bingfeng Luo, Yansong Feng, Jianbo Xu, Xiang Zhang, and Dongyan Zhao. 2017.
\newblock Learning to predict charges for criminal cases with legal basis.
\newblock In \emph{Proceedings of EMNLP}.

\bibitem[{Maxwell and Schafer(2008)}]{maxwell2008concept}
K~Tamsin Maxwell and Burkhard Schafer. 2008.
\newblock Concept and context in legal information retrieval.
\newblock In \emph{Proceedings of JURIX}.

\bibitem[{Medin(2000)}]{medin2000psychology}
Douglas~L Medin. 2000.
\newblock \emph{Psychology of learning and motivation: advances in research and
  theory}.
\newblock Elsevier.

\bibitem[{Mikolov et~al.(2013)Mikolov, Chen, Corrado, and
  Dean}]{mikolov2013efficient}
Tomas Mikolov, Kai Chen, Greg Corrado, and Jeffrey Dean. 2013.
\newblock Efficient estimation of word representations in vector space.
\newblock \emph{arXiv preprint arXiv:1301.3781}.

\bibitem[{Miwa and Bansal(2016)}]{miwa2016end}
Makoto Miwa and Mohit Bansal. 2016.
\newblock End-to-end relation extraction using lstms on sequences and tree
  structures.
\newblock In \emph{Proceedings of ACL}, pages 1105--1116.

\bibitem[{Monroy et~al.(2008)Monroy, Calvo, and Gelbukh}]{monroy2008using}
Alfredo Monroy, Hiram Calvo, and Alexander Gelbukh. 2008.
\newblock Using graphs for shallow question answering on legal documents.
\newblock In \emph{Mexican International Conference on Artificial
  Intelligence}. Springer.

\bibitem[{Monroy et~al.(2009)Monroy, Calvo, and Gelbukh}]{monroy2009nlp}
Alfredo Monroy, Hiram Calvo, and Alexander Gelbukh. 2009.
\newblock Nlp for shallow question answering of legal documents using graphs.
\newblock In \emph{Proceedings of CICLing}. Springer.

\bibitem[{Monroy et~al.(2013)Monroy, Calvo, Gelbukh, and
  Pacheco}]{monroy2013link}
Alfredo~L{\'o}pez Monroy, Hiram Calvo, Alexander Gelbukh, and
  Georgina~Garc{\'\i}a Pacheco. 2013.
\newblock Link analysis for representing and retrieving legal information.
\newblock In \emph{Proceedings of CICLing}, pages 380--393. Springer.

\bibitem[{Nagel(1963)}]{nagel1963applying}
Stuart~S Nagel. 1963.
\newblock Applying correlation analysis to case prediction.
\newblock \emph{Texas Law Review}, 42:1006.

\bibitem[{Nay(2016)}]{nay-2016-gov2vec}
John~J. Nay. 2016.
\newblock {G}ov2{V}ec: Learning distributed representations of institutions and
  their legal text.
\newblock In \emph{Proceedings of the First Workshop on {NLP} and Computational
  Social Science}.

\bibitem[{Nguyen et~al.(2016)Nguyen, Cho, and
  Grishman}]{nguyen-etal-2016-joint}
Thien~Huu Nguyen, Kyunghyun Cho, and Ralph Grishman. 2016.
\newblock Joint event extraction via recurrent neural networks.
\newblock In \emph{Proceedings of NAACL}.

\bibitem[{Nguyen and Grishman(2018)}]{nguyen2018graph}
Thien~Huu Nguyen and Ralph Grishman. 2018.
\newblock Graph convolutional networks with argument-aware pooling for event
  detection.
\newblock In \emph{Proceedings of AAAI}.

\bibitem[{Palangi et~al.(2016)Palangi, Deng, Shen, Gao, He, Chen, Song, and
  Ward}]{palangi2016deep}
Hamid Palangi, Li~Deng, Yelong Shen, Jianfeng Gao, Xiaodong He, Jianshu Chen,
  Xinying Song, and Rabab Ward. 2016.
\newblock Deep sentence embedding using long short-term memory networks:
  Analysis and application to information retrieval.
\newblock \emph{IEEE/ACM Transactions on Audio, Speech and Language Processing
  (TASLP)}, 24(4).

\bibitem[{Pan et~al.(2019)Pan, Lu, Gu, Zhang, and Xu}]{pan2019charge}
Sicheng Pan, Tun Lu, Ning Gu, Huajuan Zhang, and Chunlin Xu. 2019.
\newblock Charge prediction for multi-defendant cases with multi-scale
  attention.
\newblock In \emph{CCF Conference on Computer Supported Cooperative Work and
  Social Computing}. Springer.

\bibitem[{Pennington et~al.(2014)Pennington, Socher, and
  Manning}]{pennington2014glove}
Jeffrey Pennington, Richard Socher, and Christopher~D. Manning. 2014.
\newblock Glove: Global vectors for word representation.
\newblock In \emph{Proceedings of EMNLP}, pages 1532--1543.

\bibitem[{Peters et~al.(2018)Peters, Neumann, Iyyer, Gardner, Clark, Lee, and
  Zettlemoyer}]{peters2018deep}
Matthew~E Peters, Mark Neumann, Mohit Iyyer, Matt Gardner, Christopher Clark,
  Kenton Lee, and Luke Zettlemoyer. 2018.
\newblock Deep contextualized word representations.
\newblock \emph{arXiv preprint arXiv:1802.05365}.

\bibitem[{Peters et~al.(2019)Peters, Neumann, Logan, Schwartz, Joshi, Singh,
  and Smith}]{peters2019knowledge}
Matthew~E Peters, Mark Neumann, Robert Logan, Roy Schwartz, Vidur Joshi, Sameer
  Singh, and Noah~A Smith. 2019.
\newblock Knowledge enhanced contextual word representations.
\newblock In \emph{Proceedings of EMNLP-IJCNLP}.

\bibitem[{Qiu and Huang(2015)}]{qiu2015convolutional}
Xipeng Qiu and Xuanjing Huang. 2015.
\newblock Convolutional neural tensor network architecture for community-based
  question answering.
\newblock In \emph{Proceedings of IJCAI}.

\bibitem[{Radford et~al.(2019)Radford, Wu, Child, Luan, Amodei, and
  Sutskever}]{radford2019language}
Alec Radford, Jeffrey Wu, Rewon Child, David Luan, Dario Amodei, and Ilya
  Sutskever. 2019.
\newblock Language models are unsupervised multitask learners.
\newblock \emph{OpenAI Blog}, 1(8).

\bibitem[{Raghav et~al.(2016)Raghav, Reddy, and Reddy}]{raghav2016analyzing}
K~Raghav, P~Krishna Reddy, and V~Balakista Reddy. 2016.
\newblock Analyzing the extraction of relevant legal judgments using
  paragraph-level and citation information.
\newblock \emph{AI4JCArtificial Intelligence for Justice}, page~30.

\bibitem[{Rajpurkar et~al.(2018)Rajpurkar, Jia, and
  Liang}]{rajpurkar-etal-2018-know}
Pranav Rajpurkar, Robin Jia, and Percy Liang. 2018.
\newblock Know what you don{'}t know: Unanswerable questions for {SQ}u{AD}.
\newblock In \emph{Proceedings of ACL}.

\bibitem[{Rendle(2010)}]{rendle2010factorization}
Steffen Rendle. 2010.
\newblock Factorization machines.
\newblock In \emph{Proceedings of ICDM}. IEEE.

\bibitem[{Robertson and Walker(1994)}]{robertson1994some}
Stephen~E Robertson and Steve Walker. 1994.
\newblock Some simple effective approximations to the 2-poisson model for
  probabilistic weighted retrieval.
\newblock In \emph{Proceedings of SIGIR}.

\bibitem[{Salton and Buckley(1988)}]{salton1988term}
Gerard Salton and Christopher Buckley. 1988.
\newblock Term-weighting approaches in automatic text retrieval.
\newblock \emph{Information processing \& management}.

\bibitem[{Segal(1984)}]{segal1984predicting}
Jeffrey~A Segal. 1984.
\newblock Predicting supreme court cases probabilistically: The search and
  seizure cases, 1962-1981.
\newblock \emph{American Political Science Review}, 78(4):891--900.

\bibitem[{Seo et~al.(2016)Seo, Kembhavi, Farhadi, and
  Hajishirzi}]{seo2016bidirectional}
Minjoon Seo, Aniruddha Kembhavi, Ali Farhadi, and Hannaneh Hajishirzi. 2016.
\newblock Bidirectional attention flow for machine comprehension.
\newblock \emph{arXiv preprint arXiv:1611.01603}.

\bibitem[{Shen et~al.(2014)Shen, He, Gao, Deng, and Mesnil}]{shen2014latent}
Yelong Shen, Xiaodong He, Jianfeng Gao, Li~Deng, and Gr{\'e}goire Mesnil. 2014.
\newblock A latent semantic model with convolutional-pooling structure for
  information retrieval.
\newblock In \emph{Proceedings of CIKM}. ACM.

\bibitem[{Shu et~al.(2019)Shu, Zhao, Zeng, and Ma}]{element}
Yi~Shu, Yao Zhao, Xianghui Zeng, and Qingli Ma. 2019.
\newblock Cail2019-fe.
\newblock Technical report, Gridsum.

\bibitem[{Sugathadasa et~al.(2018)Sugathadasa, Ayesha, de~Silva, Perera,
  Jayawardana, Lakmal, and Perera}]{sugathadasa2018legal}
Keet Sugathadasa, Buddhi Ayesha, Nisansa de~Silva, Amal~Shehan Perera, Vindula
  Jayawardana, Dimuthu Lakmal, and Madhavi Perera. 2018.
\newblock Legal document retrieval using document vector embeddings and deep
  learning.
\newblock In \emph{Proceedings of SAI}. Springer.

\bibitem[{Surden(2018)}]{surden2018artificial}
Harry Surden. 2018.
\newblock Artificial intelligence and law: An overview.
\newblock \emph{Ga. St. UL Rev.}

\bibitem[{Taniguchi and Kano(2016)}]{taniguchi2016legal}
Ryosuke Taniguchi and Yoshinobu Kano. 2016.
\newblock Legal yes/no question answering system using case-role analysis.
\newblock In \emph{Proceedings of JSAI}, pages 284--298. Springer.

\bibitem[{Tran et~al.(2013)Tran, Ngo, Le~Nguyen, and
  Shimazu}]{tran2013answering}
Oanh~Thi Tran, Bach~Xuan Ngo, Minh Le~Nguyen, and Akira Shimazu. 2013.
\newblock Answering legal questions by mining reference information.
\newblock In \emph{Proceedings of JSAI}. Springer.

\bibitem[{Tran et~al.(2019)Tran, Nguyen, and Satoh}]{tran2019building}
Vu~Tran, Minh~Le Nguyen, and Ken Satoh. 2019.
\newblock Building legal case retrieval systems with lexical matching and
  summarization using a pre-trained phrase scoring model.
\newblock In \emph{Proceedings of Artificial Intelligence and Law}. ACM.

\bibitem[{Truyens and Eecke(2014)}]{truyens-eecke-2014-legal}
Maarten Truyens and Patrick~Van Eecke. 2014.
\newblock Legal aspects of text mining.
\newblock In \emph{Proceedings of LREC}.

\bibitem[{Turtle(1995)}]{turtle1995text}
Howard Turtle. 1995.
\newblock Text retrieval in the legal world.
\newblock \emph{Artificial Intelligence and Law}, 3(1-2).

\bibitem[{Ulmer(1963)}]{ulmer1963quantitative}
S~Sidney Ulmer. 1963.
\newblock Quantitative analysis of judicial processes: Some practical and
  theoretical applications.
\newblock \emph{Law and Contemporary Problems}, 28:164.

\bibitem[{Vacek et~al.(2019)Vacek, Teo, Song, Nugent, Cowling, and
  Schilder}]{vacek-etal-2019-litigation}
Thomas Vacek, Ronald Teo, Dezhao Song, Timothy Nugent, Conner Cowling, and
  Frank Schilder. 2019.
\newblock Litigation analytics: Case outcomes extracted from {US} federal court
  dockets.
\newblock In \emph{Proceedings of NLLP Workshop}.

\bibitem[{Vacek and Schilder(2017)}]{vacek2017sequence}
Tom Vacek and Frank Schilder. 2017.
\newblock A sequence approach to case outcome detection.
\newblock In \emph{Proceedings of Articial Intelligence and Law}, pages
  209--215. ACM.

\bibitem[{Van~Opijnen and Santos(2017)}]{van2017concept}
Marc Van~Opijnen and Cristiana Santos. 2017.
\newblock On the concept of relevance in legal information retrieval.
\newblock \emph{Artificial Intelligence and Law}, 25(1).

\bibitem[{Wang et~al.(2019)Wang, He, Zou, Shen, and Li}]{wang2019using}
Hui Wang, Tieke He, Zhipeng Zou, Siyuan Shen, and Yu~Li. 2019.
\newblock Using case facts to predict accusation based on deep learning.
\newblock In \emph{Proceedings of QRS-C}, pages 133--137. IEEE.

\bibitem[{Wang et~al.(2018)Wang, Yu, Jiang, and Chang}]{wang-etal-2018-co}
Shuohang Wang, Mo~Yu, Jing Jiang, and Shiyu Chang. 2018.
\newblock A co-matching model for multi-choice reading comprehension.
\newblock In \emph{Proceedings of ACL}.

\bibitem[{Wu et~al.(2013)Wu, Li, and Xu}]{wu2013learning}
Wei Wu, Hang Li, and Jun Xu. 2013.
\newblock Learning query and document similarities from click-through bipartite
  graph with metadata.
\newblock In \emph{Proceedings of WSDM}. ACM.

\bibitem[{Xiao et~al.(2018)Xiao, Zhong, Guo, Tu, Liu, Sun, Feng, Han, Hu, Wang
  et~al.}]{xiao2018cail}
Chaojun Xiao, Haoxi Zhong, Zhipeng Guo, Cunchao Tu, Zhiyuan Liu, Maosong Sun,
  Yansong Feng, Xianpei Han, Zhen Hu, Heng Wang, et~al. 2018.
\newblock Cail2018: A large-scale legal dataset for judgment prediction.
\newblock \emph{arXiv preprint arXiv:1807.02478}.

\bibitem[{Xiao et~al.(2019)Xiao, Zhong, Guo, Tu, Liu, Sun, Zhang, Han, Wang, Xu
  et~al.}]{xiao2019cail2019}
Chaojun Xiao, Haoxi Zhong, Zhipeng Guo, Cunchao Tu, Zhiyuan Liu, Maosong Sun,
  Tianyang Zhang, Xianpei Han, Heng Wang, Jianfeng Xu, et~al. 2019.
\newblock Cail2019-scm: A dataset of similar case matching in legal domain.
\newblock \emph{arXiv preprint arXiv:1911.08962}.

\bibitem[{Xu et~al.(2010)Xu, Li, and Zhong}]{xu2010relevance}
Jun Xu, Hang Li, and Chaoliang Zhong. 2010.
\newblock Relevance ranking using kernels.
\newblock In \emph{Proceedings of AIRS}. Springer.

\bibitem[{Yan et~al.(2017)Yan, Zheng, Lu, and Song}]{yan2017event}
Yukun Yan, Daqi Zheng, Zhengdong Lu, and Sen Song. 2017.
\newblock Event identification as a decision process with non-linear
  representation of text.
\newblock \emph{arXiv preprint arXiv:1710.00969}.

\bibitem[{Yang et~al.(2014)Yang, Yih, He, Gao, and Deng}]{yang2014embedding}
Bishan Yang, Wen-tau Yih, Xiaodong He, Jianfeng Gao, and Li~Deng. 2014.
\newblock Embedding entities and relations for learning and inference in
  knowledge bases.
\newblock \emph{arXiv preprint arXiv:1412.6575}.

\bibitem[{Yang et~al.(2019)Yang, Dai, Yang, Carbonell, Salakhutdinov, and
  Le}]{yang2019xlnet}
Zhilin Yang, Zihang Dai, Yiming Yang, Jaime Carbonell, Ruslan Salakhutdinov,
  and Quoc~V Le. 2019.
\newblock Xlnet: Generalized autoregressive pretraining for language
  understanding.
\newblock \emph{arXiv preprint arXiv:1906.08237}.

\bibitem[{Ye et~al.(2018)Ye, Jiang, Luo, and Chao}]{ye-etal-2018-interpretable}
Hai Ye, Xin Jiang, Zhunchen Luo, and Wenhan Chao. 2018.
\newblock Interpretable charge predictions for criminal cases: Learning to
  generate court views from fact descriptions.
\newblock In \emph{Proceedings of NAACL}.

\bibitem[{Yin et~al.(2016)Yin, Sch{\"u}tze, Xiang, and
  Zhou}]{yin-etal-2016-abcnn}
Wenpeng Yin, Hinrich Sch{\"u}tze, Bing Xiang, and Bowen Zhou. 2016.
\newblock {ABCNN}: Attention-based convolutional neural network for modeling
  sentence pairs.
\newblock \emph{Transactions of the Association for Computational Linguistics}.

\bibitem[{Yin et~al.(2018)Yin, Zheng, Lu, and Liu}]{yin2018neural}
Xiaoxiao Yin, Daqi Zheng, Zhengdong Lu, and Ruifang Liu. 2018.
\newblock Neural entity reasoner for global consistency in ner.
\newblock \emph{arXiv preprint arXiv:1810.00347}.

\bibitem[{Zeng et~al.(2015)Zeng, Liu, Chen, and Zhao}]{zeng-etal-2015-distant}
Daojian Zeng, Kang Liu, Yubo Chen, and Jun Zhao. 2015.
\newblock Distant supervision for relation extraction via piecewise
  convolutional neural networks.
\newblock In \emph{Proceedings of EMNLP}.

\bibitem[{Zhang et~al.(2017)Zhang, Pu, Yang, Zhou, and
  Gao}]{zhang2017ontological}
Ni~Zhang, Yi-Fei Pu, Sui-Quan Yang, Ji-Liu Zhou, and Jin-Kang Gao. 2017.
\newblock An ontological chinese legal consultation system.
\newblock \emph{IEEE Access}, 5:18250--18261.

\bibitem[{Zhang et~al.(2019)Zhang, Han, Liu, Jiang, Sun, and
  Liu}]{zhang-etal-2019-ernie}
Zhengyan Zhang, Xu~Han, Zhiyuan Liu, Xin Jiang, Maosong Sun, and Qun Liu. 2019.
\newblock {ERNIE}: Enhanced language representation with informative entities.
\newblock In \emph{Proceedings of ACL}.

\bibitem[{Zhong et~al.(2018)Zhong, Guo, Tu, Xiao, Liu, and
  Sun}]{zhong-etal-2018-legal}
Haoxi Zhong, Zhipeng Guo, Cunchao Tu, Chaojun Xiao, Zhiyuan Liu, and Maosong
  Sun. 2018.
\newblock Legal judgment prediction via topological learning.
\newblock In \emph{Proceedings of EMNLP}.

\bibitem[{Zhong et~al.(2020)Zhong, Wang, Tu, Zhang, Liu, and
  Sun}]{zhongiteratively}
Haoxi Zhong, Yuzhong Wang, Cunchao Tu, Tianyang Zhang, Zhiyuan Liu, and Maosong
  Sun. 2020.
\newblock Iteratively questioning and answering for interpretable legal
  judgment prediction.
\newblock In \emph{Proceedings of AAAI}.

\bibitem[{Zhong et~al.(2019{\natexlab{a}})Zhong, Xiao, Tu, Zhang, Liu, and
  Sun}]{zhong2019jec}
Haoxi Zhong, Chaojun Xiao, Cunchao Tu, Tianyang Zhang, Zhiyuan Liu, and Maosong
  Sun. 2019{\natexlab{a}}.
\newblock Jec-qa: A legal-domain question answering dataset.
\newblock \emph{arXiv preprint arXiv:1911.12011}.

\bibitem[{Zhong et~al.(2019{\natexlab{b}})Zhong, Zhang, Liu, and
  Sun}]{zhong2019openclap}
Haoxi Zhong, Zhengyan Zhang, Zhiyuan Liu, and Maosong Sun. 2019{\natexlab{b}}.
\newblock Open chinese language pre-trained model zoo.
\newblock Technical report, Technical Report. Technical Report.

\bibitem[{Zhu et~al.(2018)Zhu, Wei, Qin, and Liu}]{zhu2018hierarchical}
Haichao Zhu, Furu Wei, Bing Qin, and Ting Liu. 2018.
\newblock Hierarchical attention flow for multiple-choice reading
  comprehension.
\newblock In \emph{Proceedings of AAAI}.

\end{thebibliography}
\bibliographystyle{acl_natbib}

\end{document}


\maketitle

\appendix

\begin{table*}[]
\small
    \centering
    \begin{tabular}{c|c|c|c} \toprule
Dataset & Task & Language & Size 
\\ \midrule

\citet{chalkidis-etal-2019-large} & 
Classification & English & 57k documents, 4.3k labels
\\\midrule

\citet{cvrcek-etal-2012-legal} & 
Dictionary & Czech & 10k entries, 20k terms
\\ \midrule

\citet{element} & 
Element Extraction & Chinese & 70k sentences
\\\midrule

\citet{locke2018test} & 
Information Retrieve & English & 3m decisions, 2572 assessments
\\\midrule

\citet{kano2018coliee} &
IR and QA & Japanese & 285 queries, 651 questions
\\ \midrule

\citet{xiao2018cail} & 
Judgment Prediction &  Chinese & 2.68m documents
\\ \midrule
\citet{chalkidis-etal-2019-neural} & 
Judgment Prediction & English & 11.5k documents
\\ \midrule

\citet{hoekstra2007lkif}  & 
Ontology & English & 2378 concepts
\\\midrule

\citet{gamper-2000-parallel} & 
Parallel Corpus & Italian, German & 5m words
\\ \midrule

\citet{fawei-etal-2016-passing} & 
Question Answering & English & 400 questions
\\ \midrule

\citet{zhong2019jec} & 
Question Answering & Chinese & 30k questions, 80k articles
\\\midrule

\citet{duan2019cjrc} & 
Reading Comprehension & Chinese & 50k questions, 10k documents
\\\midrule

\citet{xiao2019cail2019} & 
Similar Case Matching & Chinese & 9k triplets of documents
\\\midrule

\citet{demenko-etal-2008-jurisdic} & 
Speech & Polish & 2h vocal material
\\ \midrule

\citet{grover-etal-2004-holj} & 
Summarization & English & 40 documents, 12k sentences
\\ \midrule

\citet{manor-li-2019-plain} & 
Summarization & English & 505 sets, 175 documents
\\

\bottomrule
    \end{tabular}
    \caption{An overview of datasets in \legal{}. The order is decided by the task and year.}
    \label{table:dataset}
\end{table*}

\section{Legal Datasets}

We show several typical datasets of \legal{}, as these datasets can be served as the benchmark of \legal{}. You can find more details from Table~\ref{table:dataset}.

\bibliography{anthology,acl2020}
\bibliographystyle{acl_natbib}